\pdfoutput=1

\documentclass[11pt]{article}

\usepackage{acl}

\usepackage{times}
\usepackage{latexsym}

\usepackage[T1]{fontenc}

\usepackage[utf8]{inputenc}

\usepackage{microtype}

\usepackage{amsmath}
\usepackage{amsfonts}
\usepackage{algorithm}
\usepackage{multirow}
\usepackage{graphicx}
\usepackage{algorithmic}
\usepackage{microtype}
\usepackage{booktabs} 
\usepackage{subfigure}
\usepackage{bm}
\usepackage{amsfonts}
\usepackage{soul}
\usepackage{makecell}
\usepackage{microtype}
\usepackage{booktabs} 
\usepackage{subfigure}
\usepackage{bm}
\usepackage{amsfonts}
\usepackage{soul}
\setul{0.5ex}{0.3ex}
\setulcolor{blue}
\usepackage{makecell}
\usepackage{nicefrac}       
\usepackage{booktabs,multirow}

\usepackage{microtype} 
\usepackage{epsfig}
\usepackage{diagbox}

\usepackage{framed}
\usepackage{empheq}
\usepackage{array}
\usepackage{enumitem}
\usepackage{mdwlist}
\usepackage{cleveref}

\newcommand{\thickhline}{\noalign{\hrule height 1pt}}

\title{There Are a Thousand Hamlets in a Thousand People's Eyes: \\
Enhancing Knowledge-grounded Dialogue with Personal Memory}

\author{\\
	\textbf{Tingchen Fu\textsuperscript{1}\footnotemark[2],  Xueliang Zhao\textsuperscript{2}\footnotemark[2], Chongyang Tao\textsuperscript{3}, Ji-Rong Wen\textsuperscript{1}, Rui Yan\textsuperscript{1}\footnotemark[1]}\\
	\textsuperscript{1}Gaoling School of Artificial Intelligence, Renmin University of China \\
	\textsuperscript{2}Wangxuan Institute of Computer Technology, Peking University\\
	\textsuperscript{3}Microsoft Corporation\\
	\texttt{\{lucas.futingchen,zhaoxlpku,chongyangtao\}@gmail.com}\\ \texttt{\{jrwen,ruiyan\}@ruc.edu.cn}
}

\begin{document}
\maketitle

\renewcommand{\thefootnote}{\fnsymbol{footnote}}
\footnotetext[2]{The first two authors contribute equally. Xueliang Zhao is responsible for the design of the methodology and algorithm. Tingchen Fu is responsible for the implementation and experiment. The order is decided by a coin flip.}
\footnotetext[1]{Corresponding author: Rui Yan (ruiyan@ruc.edu.cn)}
\setcounter{footnote}{0}
\renewcommand{\thefootnote}{\arabic{footnote}}

\begin{abstract}
\label{sec:abstract}

Knowledge-grounded conversation~(KGC) shows great potential in building an engaging and knowledgeable chatbot, and knowledge selection is a key ingredient in it. 
However, previous methods for knowledge selection only concentrate on the relevance between knowledge and dialogue context, ignoring the fact that age, hobby, education and life experience of an interlocutor have a major effect on his or her personal preference over external knowledge. Without taking the personalization issue into account, it is difficult to select the proper knowledge and generate persona-consistent responses.
In this work, we introduce personal memory into knowledge selection in KGC to address the personalization issue. We propose a variational method to model the underlying relationship between one's personal memory and his or her selection of knowledge, and devise a learning scheme in which the forward mapping from personal memory to knowledge and its inverse mapping is included in a closed loop so that they could teach each other.  
Experiment results show that our method outperforms existing KGC methods significantly on both automatic evaluation and human evaluation.

\end{abstract}

\section{Introduction}
\label{sec:intro}
Open-domain dialogue system often suffers from safe response~\cite{li2015diversity,zhang2019dialogpt} problem as they could only refer to the context when generating a response. To alleviate this, knowledge-grounded conversation~(KGC) is proposed to introduce external fact and real-world commonsense as prior knowledge~\cite{zhou2018commonsense,dinan2018wizard,zhao2020low}, such that a dialogue system is able to ground the conversation with the provided knowledge and therefore generate informative and engaging responses. 
As external knowledge supplements the background to the inputs and decides what to say, knowledge selection is a key ingredient in KGC.

Numerous methods have been developed to tackle the knowledge selection problem by sequential latent variables~\cite{kim2020sequential,meng2020dukenet}, reinforcement learning~\cite{zhao2020knowledge}, or expectation maximization algorithm~\cite{li2020zero}. In spite of the progress in this task, knowledge selection remains an unsolved problem as the precision is still far from satisfactory in Wizard of Wikipedia~\cite{dinan2018wizard} and other benchmarks in KGC~\cite{gopalakrishnan2019topical}, {which also hinders the optimization of subsequent response generation models.} 
A crucial point is, they often make assumption that the golden knowledge is distinguishable as long as the dialogue context is known, yet this is not always held true because there exists a one-to-many relationship in conversation and the past utterance history in a dialogue session is insufficient to decide the knowledge selection or the future trend of a dialogue.

\begin{figure}
    \centering
    \includegraphics[width=0.98\linewidth]{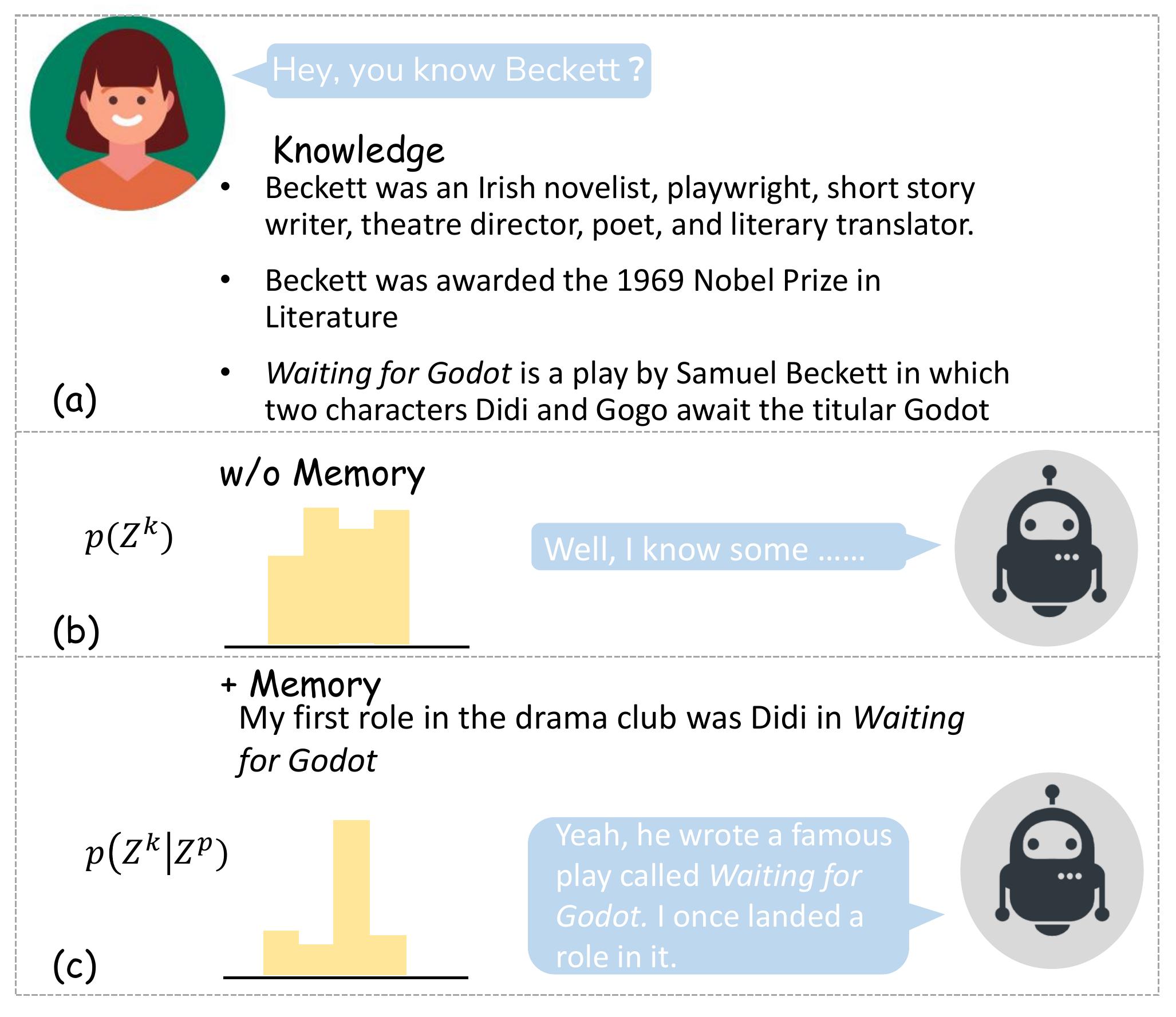}
    \vspace{-2.5mm}
    \caption{(a)The knowledge selection could not be certainly determined only based on dialogue context. (b) With out personal memory, the knowledge probability distribution is flat and is difficult to choose the proper knowledge. (c) Enhanced with personal memory, the knowledge probability distribution is sharper.}
    
    \label{fig:intro_case}
\vspace{-7mm}
\end{figure}

As is shown in Figure~\ref{fig:intro_case}, personalization is a key to success in the task because knowledge selection is a personal or subjective process in nature. When people communicate with each other, their perception of dialogue context will evoke their past memory about relevant life experience, taste and values, which we refer to as \textit{personal memory}. The aroused fragment of personal memory further guides their interest and preference for different knowledge.
Motivated by this, we postulate a new task named personalized KGC, introducing personalization into knowledge-grounded dialogue to encourage more human-like knowledge selection.

Importing persona memory into knowledge selection is a non-trivial task. One of the challenge is concretization of personal memory. 
Personal memory is an abstract concept related to user-specific experience, which is difficult to depict or model. 
Though it has been discussed in open-domain dialogue~\cite{li2016deep,zhang2018personalizing}, no previous research sheds light on the personalization issue in KGC and there exists no dialogue dataset featured with external facts and personal memory at the same time.
Besides, there is no annotated label to indicate which knowledge candidate a person will choose based on his or her personal memory. Namely, the mapping between personal memory and knowledge selection is highly unconstrained without golden label. Intuitive resolution like treating personal memory as additional knowledge is sub-optimal because of dependency between knowledge and personal memory, as is shown in our experiments.

To address the above issue, we construct a KGC dataset featured with personalized memory repository, collecting user-specific utterance history under multiple types of context, which is a reflection of one's personal memory. And to discover the underlying relationship between the dialogue context, personal memory and knowledge, we propose a variational method and introduce two latent variables $Z^p$ and $Z^k$ to indicate the fragment of personal memory to evoke and the knowledge candidate to select respectively. 
And to model the mapping from $Z^p$ to $Z^k$, we introduce an inverse mapping as a dual task and employ dual learning to allow the two mappings to teach each other. The motivation behind this is intuitive: The reconstruction of personal memory from selected knowledge candidate is natural and easy if the mapping from personal memory to knowledge is accurate. Extensive experiment shows that our methods outperform competitive baselines in both automatic evaluation and human evaluation, justifying the importance of introducing personal memory and the effect of the dual learning mechanism empirically.

The contributions of this work are three-fold:

(1) We explore the personalization issue of the knowledge selection task in KGC and construct a dataset featured with user-specific personal memory to benefit relevant research in the future. We are the first to explore the possibility of introducing personal memory into KGC.

(2) We propose a novel variational method and introduce two latent variables to model the inter-dependency between the persona and knowledge. Besides, we employ dual learning to optimize the relationship between the dialogue context, personal memory and knowledge in a unified framework.

(3) We conduct extensive experiments and verify the proposed methods empirically. Both the automatic and human evaluation evidence the efficacy of our proposed method.
\section{Related Work}
There is a substantial literature in the field of knowledge-grounded conversation. With the grounding of external knowledge in format of knowledge graph~\cite{zhou2018commonsense,wu2019proactive}, document~\cite{ghazvininejad2018knowledge,zhou2018dataset,zhao2019document} or visual background~\cite{das2017visual}, it is regarded as a critical method towards intelligent dialogue system. Nowadays, existing methods in KGC often share a paradigm that decomposes the task into two related sub-problems, namely knowledge selection and utterance generation~\cite{kim2020sequential}. In this work, we mainly focus on the knowledge selection task. 
To this end, a great deal of methods have been proposed to retrieve the most relevant knowledge by memory network~\cite{ghazvininejad2018knowledge}, sequential latent variables~\cite{kim2020sequential,meng2020dukenet}, reinforcement learning~\cite{zhao2020knowledge} and so on. A recent work gives attention to the expression style of knowledge~\cite{zhao2021learning}. However, they only focus on the decoding phase and no methods shed light on the personalization issue of knowledge selection, to our best knowledge.

Our work is related to dual learning as well. First proposed in neural machine translation by \citet{he2016dual}, dual learning is a semi-supervision learning scheme aiming at utilizing large-scale unlabeled data. 
Together with its newly appeared variants in recent years~\cite{xia2017dual,xia2018model,wang2019multi}, dual learning has been successfully applied in neural machine translation~\cite{xia2017dual,he2017decoding},  image-image-translation~\cite{DBLP:conf/iccv/YiZTG17,lin2018conditional}, sentiment analysis~\cite{xia2017dual}, automatic speech recognition~\cite{ren2019almost}, question answering~\cite{tang2017question}, 
and knowledge-grounded dialogue~\cite{meng2020dukenet}. Our work is related to dual learning as well. First proposed in neural machine translation by \citet{he2016dual}, dual learning is a semi-supervision learning scheme aiming at utilizing the large scale unlabeled data.
In this work, we apply dual learning to model the inter-dependency relationship between one's personal memory and his or her choice of knowledge.
\section{Methodology}
\label{sec: methodology}

\subsection{Problem Formulation}
\label{sec: formulation}
Suppose we have a KGC dataset $\mathcal{D}$ with $N$ case, and every case is in format of $(C,\mathcal{K},R)$, where $C=[u_1,u_2,\cdots,u_{l_C}]$ is the context of the dialogue with $l_C$ tokens in total,
$\mathcal{K}=\{K_1,K_2,\cdots,K_{|\mathcal{K}|}\}$ is a set of $|\mathcal{K}|$ knowledge candidates. And $R=[r_1,r_2,\cdots,r_{l_R}]$ is a response in this conversation corresponding to a specific user with unique user id. Different from the original KGC task, we have a memory repository $\mathcal{M}$. For every interlocutor corresponding to the response, a set of his or her personal memory $\mathcal{P}=\{P_1,P_2,\cdots,P_{|\mathcal{P}|}\}$ composed of $|\mathcal{P}|$ customized utterance history could be retrieved from the memory repository.   
Our goal is to learn a probabilistic model $p(R|C,\mathcal{K},\mathcal{P})$ that could generate a personalized and informative response based on personal memory and knowledge.

\subsection{Model Overview}
\label{sec: overview}
\begin{figure}
    \centering
    \includegraphics[width=1.0\linewidth]{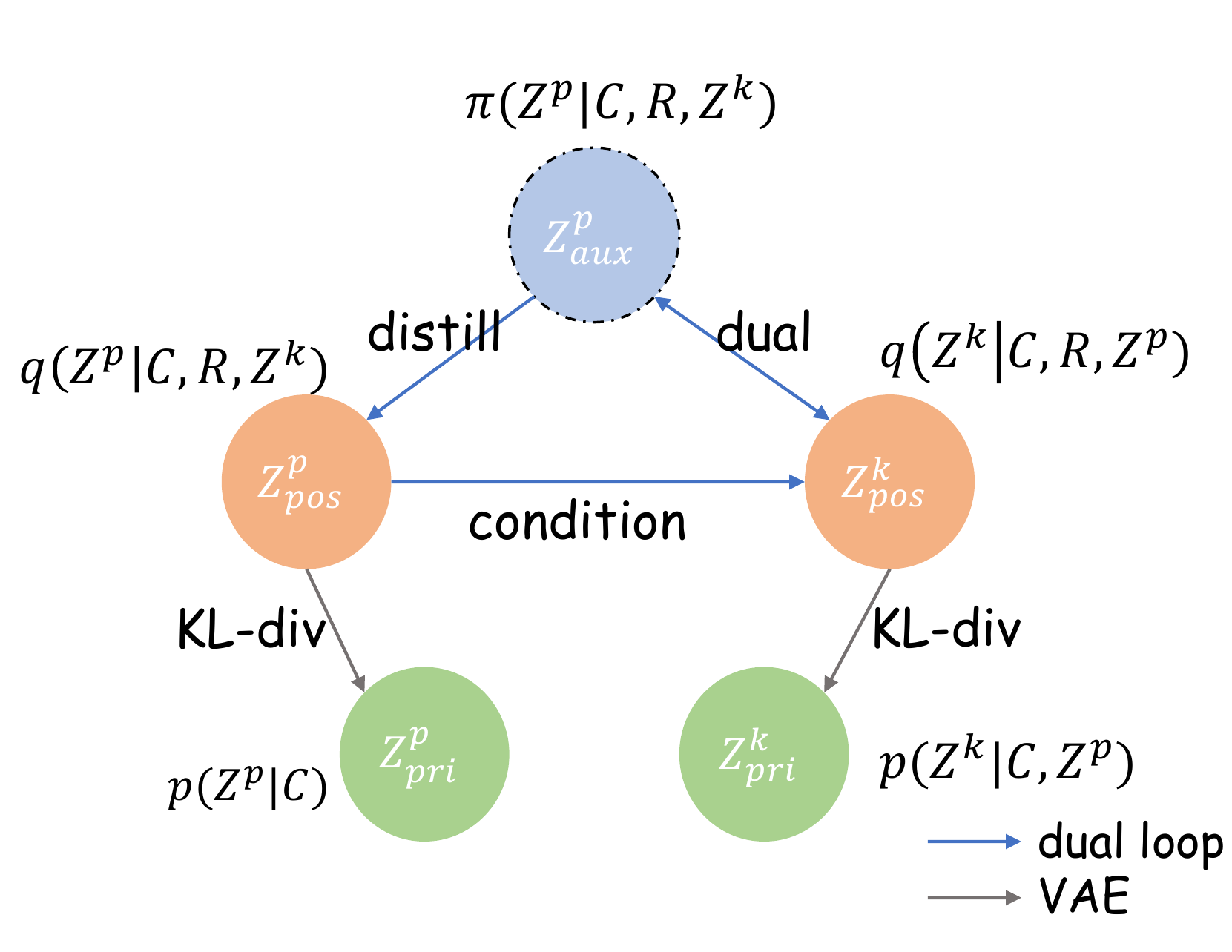}
    \vspace{-5mm}
    \caption{A graphical representation of our proposed method. It depicts the dependency and interaction between $Z^p$ and $Z^k$.}
    \label{fig:graph model}
    \vspace{-6mm}
\end{figure}

Figure~\ref{fig:graph model} gives a graphical model of our methods. As is shown, the core of our proposed method is five probabilistic models to calculate the prior and posterior distribution of $Z^p$, $Z^k$ and an auxiliary distribution of $Z^p$.  
During training, we devise an unsupervised learning scheme, in which we optimize the distribution of two latent variables $Z^p$ and $Z^k$ by dual learning. To be more specific, we first sample a $\tilde{Z^p}$ from the posterior distribution $q_{\phi}(Z^p|C,R)$, and then calculate the forward mapping from memory to knowledge $q_\phi(Z^k|C,R,\tilde{Z^p})$, from which we sample a $\tilde{Z^k}$.
The reward is designed as the probability of reconstructing the selected memory fragment by the auxiliary distribution $\pi_{\psi}(Z^p=\tilde{Z^p}|C,R,\tilde{Z^k})$. By maximizing the reward, the primal task and the auxiliary task could benefit each other.
The gains of the auxiliary distribution is distilled to $q_\phi(Z^p|C,R)$, such that the two posterior distribution and the auxiliary distribution form a closed loop. Besides, the prior distribution is forced to get close to the posterior distribution via KL-divergence.

In the inference phase, the prior distribution of $Z^p$ is calculated at first, from which we sample and activate a personal memory fragment. After that, the woken memory fragment is used to decide the prior knowledge distribution $p_\theta(Z^k|C)$. Finally, the knowledge sampled from $Z^k$ together with the memory fragment is sent into a generator to synthesize a response. Note that the golden response is only involved in the training phase. $\pi$, $\phi$ and $\psi$ are all learnable parameters.

\subsection{Neural parameterization}
\label{sec: neural}
To make the latent variables interpretable, we set the latent space of $Z^p$ and $Z^k$ as the number of memory fragments or knowledge candidates to choose from, and each sampling corresponds to a single piece of memory fragment or a knowledge candidate.
Furthermore, motivated by human cognitive process, the aroused personal memory fragment implies one's preference for different external knowledge, which influences the likelihood of choosing different knowledge. 
In light of this, the prior distribution of $(Z^p,Z^k)$ is factorized as:
\begin{equation}
    p(Z^p,Z^k)=p(Z^k|Z^p)p(Z^p)
\end{equation}
And to calculate their probability distribution, we adopt BERT~\cite{devlin2018bert} as the backbone of our method to obtain a dense representation of dialogue context, response, candidate knowledge sentence or personal memory fragment. 
Take the calculation of the prior distribution $p_{\theta}(Z^k|C,Z^p)$ as an example. 
We first concatenate the context $C$, the memory fragment $P$ indicated by the sampled $Z^p$, and the $i$-th candidate knowledge $K_i$ together as a long sequence. A special [CLS] token is prepended at the beginning of the sequence and [SEP] is inserted to separate different utterances:
\begin{equation}
\begin{small}
\begin{aligned}
&\mathcal{I}=u_1,u_2,\cdots u_{l_C},p_1, p_2,\cdots,p_{l_P},k_1,k_2,\cdots k_{l_{K_i}}, \\
\end{aligned}
\end{small}
\end{equation}
where $l_C$, $l_P$ and $l_{K_i}$ are the number of tokens in the context, memory facet and knowledge candidate respectively. Then the embedding layer will convert $\mathcal{I}$ into input representations, which is the sum of the corresponding token embedding and position embedding. Thereafter, the BERT encoder performs multi-head attention on the input representation to obtain a dense representation. There are $n$ identical layers in the BERT encoder, and for each layer, the multi-head attention could be formulated as 
\begin{equation} \small
    \mathbf{H}^{l}=\mathrm{FFN}(\mathrm{MultiHead}(\mathbf{Q}^{l-1},\mathbf{K}^{l-1},\mathbf{V}^{l-1})),
\end{equation}
where $\mathrm{FFN}(\cdot)$ is a feed-forward network and we use $\mathbf{Q}^{l-1}$, $\mathbf{K}^{l-1}$, and $\mathbf{V}^{l-1}$ to denote the query matrix, key matrix and value matrix after the $l-1$-th layer respectively. For self-attention, we have 
\begin{equation}
    \mathbf{Q}^{l-1}=\mathbf{K}^{l-1}=\mathbf{V}^{l-1}=\mathbf{H}^{l-1},
\end{equation}
where $\mathbf{H}^l$ means the hidden state at the $l$-th layer. Specially, $\mathbf{H}^0$ is the input embedding and $\mathbf{H}^n$ is the final output of the BERT.

We use the vector corresponding to the position of the special [CLS] token in $\mathbf{H}^n$ as the representation of the $i$-th knowledge candidate, which is referred to as $\mathbf{h}_i$. 
Then the distribution of $Z^k$ is calculated as 
\begin{equation}
    \begin{aligned}
        p_\theta(Z^k=i|C,Z^p)=\frac{\exp(f(\mathbf{h}_i))}{\sum\limits_{j}^{|\mathcal{K}|}\exp(f(\mathbf{h}_j))},
    \end{aligned}
\end{equation}
where $f(\cdot)$ is a multi-layer perceptron. The prior and posterior distribution of $Z^k$ and $Z^p$ are calculated in a similar way. The only difference lies in the constitution of input sequence $\mathcal{I}$: For the prior distribution of $Z^p$, $\mathcal{I}$ is the concatenation of dialogue context and a candidate personal memory facet:
\begin{equation}
    \mathcal{I}=u_1,u_2,\cdots u_{l_C},p_1, p_2,\cdots,p_{l_P}
\end{equation}
And to calculate the posterior distribution, we insert the response tokens behind the dialogue context tokens as the response usually contains clue indicating the selected knowledge and memory. Namely, to compute $q_\phi(Z^p|C,R)$, the posterior of $Z^p$, the input is:
\begin{equation}
\mathcal{I}=u_1,u_2,\cdots u_{l_C}, r_1,r_2,\cdots,r_{l_R},p_1, p_2,\cdots,p_{l_P}
\end{equation}
And for $q_\phi(Z^k|C,R,Z^p)$:
\begin{equation}
\begin{aligned}
\mathcal{I}&=u_1,u_2,\cdots u_{l_C}, r_1,r_2,\cdots,r_{l_R},\\
&p_1, p_2,\cdots,p_{l_P},k_1,k_2,\cdots,k_{l_K}
\end{aligned}
\end{equation}

Normally, the generator $g$ of our method could be specified as any large-scale pre-trained language model. Here we define the generator as GPT-2~\cite{radford2019language}. Previous methods often synthesize a response merely based on the dialogue context and the selected knowledge, taking no consideration of the persona of the interlocutor, which may lead to an inconsistency in persona. Different from that, we input the sampled personal memory fragment and the sampled knowledge candidate into GPT-2 all together with the dialogue context. Intuitively, personal memory fragment implies why the knowledge is paid attention to and underlying relevance between the persona of the interlocutor and the knowledge, which endows the generator to generate persona-consistent and knowledgeable responses: 

\begin{equation}
\begin{aligned}
 g(R)&=g(R|C,Z^p,Z^k) \\
 &=\prod\limits_{i=1}^{l_R}g(r_i|C,Z^p,Z^k,r_{<i})
\end{aligned}
\end{equation}
 
\subsection{Learning Details}
Directly maximizing the marginal log-likelihood of generating the correct response $g(R|C,Z^p,Z^k)$ requires integrating over all possibilities of $Z^k$ and $Z^p$, which is more than time-consuming. Inspired by variational inference, we introduce a variational posterior as the true posterior is intractable. Thereby, instead of directly optimizing the marginal log-likelihood, we derive an evidence lower bound objective to maximize:
\begin{equation} \small
\begin{aligned}
\label{eq:vae}
    \mathcal{L}_{ELBO}&=\mathbb{E}_{q_{\phi}(Z^k|Z^p)q_\phi(Z^p)}g(R|C,Z^p,Z^k)\\
    &-\mathbb{E}_{q_\phi(Z^p)}KL(q_\phi(Z^k|Z^p)||p_\theta(Z^k|Z^p))\\
    &-KL(q_\phi(Z^p)||p_\theta(Z^p))
\end{aligned}
\end{equation}
where $q_{\phi}(Z^k|Z^p)$, $q_\phi(Z^p)$, $p_\theta(Z^p)$, $p_\theta(Z^k|Z^p)$ are shorthand for $q_{\phi}(Z^k|C,R,Z^p)$, $q_\phi(Z^p|C,R)$, $p_\theta(Z^p)$ and $p_\theta(Z^k|C,Z^p)$ respectively.
A step-wise derivation could be found in the supplementary materials.

The forward mapping from personal memory to knowledge candidates is relatively implicit and obscure, partially because the customized utterance history contains unwanted noise. As a result, there is a tendency that $Z^p$ is ignored and $p_\theta(Z^k|Z^p,C)$ is degenerated into $p_\theta(Z^k|C)$, which we refer to as the \textit{vanishing memory}.

\begin{algorithm}[!t]
\small
\begin{algorithmic}[1]
\STATE {\bfseries Input:} Training KGC dataset $\mathcal{D}$, memory repository $\mathcal{M}$
\STATE Warm up $p_\theta(Z^p)$, $p_\theta(Z^K|Z^p)$, $q_\phi(Z^p|R)$ and $q_\phi(Z^k|R,Z^p)$ on $\mathcal{D}$. 
\WHILE{not converge} 
    \STATE Sample a mini-batch $\{(C, \mathcal{K}, R)\}$ from $\mathcal{D}$.
    \STATE Retrieve the user-specific personal memory $\mathcal{P}$ from the memory repository.
    \STATE Calculate the prior personal memory distribution $p_\theta(Z^p)$ with $C$.
    \STATE Sample a $Z^p$ and then calculate the prior distribution of knowledge $p_\theta(Z^k|Z^p)$.
    \STATE Calculate the posterior memory distribution $q_\phi(Z^p|R)$ based on $C$ and $R$, and then sample a $\tilde{Z^p}$ from that. 
    \STATE Calculate the posterior knowledge distribution $q_\phi(Z^k|R,\tilde{Z^p})$, and then sample a $\tilde{Z^k}$ from that.
    \COMMENT{The primal task}
    \STATE Compute the reward $Re_1$ as the
    Reconstruct probability $\pi_\psi(Z^p=\tilde{Z^p}|Z^k)$. 
    \STATE Update $\phi$ according to Eq.~\ref{eq:update phi}.
    \STATE Calculate the auxiliary memory distribution $\pi_\psi(Z^p|R,\bar{Z^k})$ based on the pseudo knowledge label $\bar{Z^k}$, and sample a $\tilde{Z^p}$ from that.\COMMENT{The dual task}
    \STATE Compute the reward $Re_2$ as $q_\phi(Z^k=\bar{Z^k}|\tilde{Z^p})$.
    \STATE Update $\psi$ according to Eq.~\ref{eq:upadte psi}.
    \STATE Update $\theta$ according to Eq.~\ref{eq:vae}.
    \STATE Update $\phi$ according to Eq.~\ref{eq:distill}.
\ENDWHILE
\STATE {\bfseries return} The prior distribution $p_\theta(Z^p)$ and $p_\theta(Z^K|Z^p)$
\end{algorithmic}
\caption{The proposed learning algorithm.}
\label{alg:alg}
\end{algorithm}

To address this issue, inspired by the idea of dual learning~\cite{he2016dual}, we introduce an inverse mapping from knowledge candidate to personal memory as a dual task, which is depicted by the auxiliary distribution $\pi_\psi(Z^p|C,R,Z^k)$. Intuitively, there is a natural duality between the mapping from personal memory to knowledge and the inverse mapping. Therefore, if the forward mapping makes a good inference about the knowledge to choose, the inverse mapping is able to map it back to  personal memory, which means that the memory is not vanishing.

And before the dual learning procedure, the primal task and the dual task are warmed up to speed up convergence and alleviate error accumulation in the dual learning process, following the idea of \citet{he2016dual} and \citet{meng2020dukenet}. 
Namely, we construct pseudo knowledge label $\bar{P}$ and persona label $\bar{K}$ based on their similarity to the response.
\begin{equation}
\begin{aligned}
    &\bar{K}=\mathop{\max}_{K_i\in \mathcal{K}} Sim(K_i,R)\\
    &\bar{P}=\mathop{\max}_{P_i \in \mathcal{P} } Sim(P_i,R)\\ 
    \label{eq:pseudo}
\end{aligned}
\end{equation}
Then, both the primal task and the dual task are warmed up with a traditional maximum likelihood estimation objective.

After the warm-up procedure, for each iteration, we first sample a $\tilde{Z^p}$ according to its posterior distribution $q_\phi(Z^p|C,R)$. Then the forward mapping calculates the probability distribution $q_\phi(Z^k|C,R,\tilde{Z^p})$, from which we sample a $\tilde{Z^k}$. The reward for the forward mapping is defined as the probability that the auxiliary distribution recovers the $\tilde{Z^p}$. Mathematically, we have 
\begin{equation}
    Re_1=\pi_\psi(Z^p=\tilde{Z^p}|C,R,\tilde{Z^k})
\end{equation}
Symmetrically, the reward for the auxiliary distribution is the prediction probability of the golden knowledge by the forward mapping:
\begin{equation}
    Re_2=q_\phi(Z^k=\bar{Z^k}|C,R,Z^p),
\end{equation}
where $\bar{Z^k}$ is corresponding to the pseudo knowledge label.

And the objective of the dual learning is to maximize the reward:
\begin{equation}
    \mathcal{L}_{dual}=\mathbb{E}_{\mathcal{D}}[Re_1+Re_2]
\end{equation}

For reward maximization, we optimize the parameter through policy gradient method~\cite{sutton2000policy}:
{\small
\begin{align}  
    \label{eq:upadte psi}
    \nabla_{\psi}\mathcal{L}_{dual}=\nabla_{\psi}\log \pi_\psi(Z^p=\tilde{Z^p}|C,R,\bar{Z^k}) Re_2. \\
    \nabla_{\phi}\mathcal{L}_{dual}=\nabla_{\phi}\log q_\phi(Z^k=\tilde{Z^k}|C,R,\tilde{Z^p}) Re_1.
    \label{eq:update phi}
\end{align} 
}
Finally, the gains of the dual task is distilled into the posterior distribution of $Z^p$ via a cross-entropy loss:
\begin{equation} \small
\label{eq:distill}
\begin{aligned}
    \mathcal{L}_{dis}&=-KL(\pi^T_\psi(Z^p|C,R,Z^k)||q^T_\phi(Z^k|C,R,Z^p)) \\
    &+ \alpha\log q_\phi(Z^p=\bar{Z^p}|C,R,Z^k),
\end{aligned}
\end{equation}
where $\alpha$ is a hyper-parameters to balance the weights of two parts and the superscript $T$ means that the distribution is normalized at temperature $T$.
Thus, the three probabilistic models form a closed loop in which each component is trained alternatively. 
The full procedure of our proposed learning algorithm is concluded in Algorithm~\ref{alg:alg}.

\section{Experiment}
\label{sec:exp}
\subsection{Dataset}
\label{sec:dataset}
Since existing dataset like CMU\_DoG~\cite{zhou2018dataset} or Holl-E~\cite{moghe2018towards} do not contain information about personal memory, we establish a new KGC dataset equipped with a memory repository. 
The dataset is constructed based on Reddit~\cite{baumgartner2020pushshift}. 

In detail, we download the conversational data on the PushShift dump of Reddit ranging from 2011 to the first half of 2015 and divide them into a training set, a validation set and a test set according to the date.
To construct a memory repository, we maintain a dictionary where the key is a long string hashed from the user account name and the value is a set of utterances of the user. Since it is a repository for user-specific utterances, it may inevitably contain false beliefs or subjective opinions. We shall leave this issue for future work.
Elaborated data filtering is conducted to ensure: (1) We only keep utterances from users that have at least $5$ utterances in the memory repository; (2) Utterances that are too long or too short are filtered;  (3) Paraphrase tool~\cite{prithivida2021parrot} is applied on every utterances to avoid tracing the utterances back to real reddit users. 

The statistics of our dataset is shown in Table~\ref{tab:dataset}. And the code is available at \url{https://github.com/Lucasftc/PersonaKGC}. A few examples is shown in Appendix~\ref{sec:example}. To benefit future research and meanwhile avoid possible malicious abuse, the dataset is available upon request from the authors\footnote{Please contact lucas.futingchen@gmail.com for the dataset.}.

\subsection{Compared Methods}
To verify the effectiveness of the proposed methods, we compare our methods with baselines in KGC. Meanwhile, since our proposed method makes use of personal memory to generate persona-consistency response, we also compare our methods with baselines in personalized dialogue.
\begin{itemize}[noitemsep,topsep=0.4pt,parsep=0pt,partopsep=0pt]
    \item \emph{Generative Profile Memory Network (GPMN)} \cite{zhang2018personalizing} is a method in personalized dialogue 
    which employs Memory Network along with persona information.
    \item \emph{Transformer Memory Network~(TMN)}~\cite{dinan2018wizard} adopts the traditional Memory Network with transformer architecture and introduces the knowledge selection loss.
    \item \emph{Transfertransfo}~\cite{wolf2019transfertransfo} is a combination of a transfer learning based training scheme and a high-capacity transformer model and achieves the best results in the Conversational Intelligence Challenge 2.
    \item \emph{Sequential Knowledge Transformer~(SKT)} \cite{kim2020sequential} utilizes sequential latent variables for knowledge selection. We use the pseudo knowledge labels for the golden knowledge label in implementation.
    \item \emph{KnowledGPT} \cite{zhao2020knowledge} puts the knowledge selector and the response generator in a framework and employ reinforcement learning and curriculum learning to accomplish the state-of-the-art performance in KGC.
    \item \emph{KnowledGPT+M}, a variant of KnowledGPT where we treat personal memory as knowledge candidates as well and input them to the knowledge selector.
    \item \emph{$\mathcal{P}^2$BOT} \cite{liu2020you} is a transmitter-receiver based framework explicitly modeling the perception between the interlocutors and achieves the state-of-the-art in personalized dialogue.
    \item  \emph{BoB} \cite{song-etal-2021-bob} is a newly published method that disentangles personalized dialogue into persona understanding and personalized generation. 
\end{itemize}
For more implementation details about the baselines and our method, please refer to appendix~\ref{sec:impl}. 

\begin{table}[!t]
\resizebox{1.0\linewidth}{!}{
\begin{tabular}{lccc}
\toprule
                 & Train   & Valid  & Test   \\ \midrule
\# Dialogues     & 217,095 & 11,186 & 6,236   \\
\# Utterances    & 1,442,975 & 74,480  & 41,519  \\
\# Knowledges    & 5,459,744 & 290,349 & 148,057 \\
\# User          & 48,858   & 5,603   & 3,281   \\
\# Memory facets & 490,460  & 70,494  & 38,354  \\ \midrule
\multicolumn{4}{l}{AvG.Len (\# words):}      \\
Utterance       & 34.15   & 33.95  & 33.60  \\
Knowledge        & 54.54   & 52.39  & 53.17  \\
Memory facet    & 42.10   & 40.21  & 40.60  \\ \bottomrule
\end{tabular}
}
\vspace{-2mm}
\caption{The statistics of the dataset.}
\vspace{-4mm}
\label{tab:dataset}
\end{table}

\begin{table*}[!t]
\centering \small
\resizebox{0.9\linewidth}{!}{
\begin{tabular}{lcccccccccc}
\toprule
\multirow{2}{*}{Models}    & \multicolumn{4}{c}{BLEU}   & \multicolumn{3}{c}{ROUGE} & \multicolumn{2}{c}{Distinct} & \multicolumn{1}{c}{ \multirow{2}{*}{METEOR}} \\
                \cmidrule(lr){2-5} \cmidrule(lr){6-8}  \cmidrule(lr){9-10} 
                & B-1   & B-2  & B-3  & B-4  & R-1     & R-2    & R-3    & D-1          & D-2           & \multicolumn{1}{c}{}                        \\ \midrule
                
GPMN            & 3.87  & 1.41 & 0.43 & 0.13 & 4.25    & 0.23   & 3.94   & 0.06         & 0.15          & 2.30                                        \\
TMN             & 1.05  & 0.31 & 0.12 & 0.02 & 8.91    & 1.38   & 7.88   & 0.10         & 0.28          & 2.60                                        \\
Transfertransfo & 6.09  & 1.57 & 0.62 & 0.34 & 9.31    & 0.73   & 7.34   & 8.33         & 43.43         & 3.79                                        \\
SKT             & 3.48  & 0.85 & 0.28 & 0.10 & 7.95    & 0.94   & 6.95   & 3.41         & 14.35         & 2.75                                        \\
KnowledGPT      & 5.22  & 1.76 & 0.77 & 0.39 & 10.68   & 1.71   & 9.12   & 6.65         & 28.64         & 4.09                                        \\
KnowledGPT+M    & 7.81  & 3.55 & 2.46 & 2.02 & 10.79   & 2.82   & 9.32   & 7.37         & 35.13         & 4.58                                        \\
P2bot           & 5.95  & 1.61 & 0.57 & 0.24 & 7.54    & 0.72   & 6.54   & 4.98         & 17.74         & 3.20                                        \\
BoB             & 4.69  & 1.57 & 0.65 & 0.31 & 10.68   & 1.57   & 9.30   & 4.94         & 17.06         & 3.97                                        \\ \midrule
Ours            & \textbf{13.09} & \textbf{6.22} & \textbf{4.23} & \textbf{3.33} & \textbf{13.60}   & \textbf{3.73}   & \textbf{10.64}   & \textbf{8.97}         & 39.29         & \textbf{6.65}         \\
\bottomrule
\end{tabular}
}
\vspace{-1mm}
\caption{Automatic evaluation results. Numbers in bold mean that the improvement to the best performing baseline is statistically significant (t-test with $p$-value $<$ 0.05).}
\vspace{-4mm}
\label{tab:main_exp}
\end{table*}

\begin{table}[!t]
\centering
\resizebox{1.0\linewidth}{!}{
\begin{tabular}{ccccc}
\toprule
                & Fluency & Coherence & Faithfulness &Kappa\\ \midrule
Transfertransfo & 1.65    & 1.73      & 1.68         &0.72\\
KnowledGPT+M    & 1.71    & 1.72      & 1.77         &0.69\\
BoB             & 1.67    & 1.62      & 1.70         &0.70\\
Ours            & \textbf{1.77}    & \textbf{1.79} & \textbf{1.82} &0.69        \\
\bottomrule
\end{tabular}
}
\vspace{-2mm}
\caption{Human evaluation result. Numbers in bold mean that the improvement to the best performing baseline is statistically significant (t-test with $p$-value $<$ 0.05).}
\vspace{-5mm}
\label{tab: human}
\end{table}

\begin{table*}[!t]
\centering \small
\begin{tabular}{ccccccccccc}
\toprule
\multirow{2}{*}{Models} & \multicolumn{4}{c}{BLEU} & \multicolumn{3}{c}{ROUGE} & \multicolumn{2}{c}{Distinct} & \multicolumn{1}{c}{\multirow{2}{*}{METEOR}} \\ 
\cmidrule(lr){2-5} \cmidrule(lr){6-8}  \cmidrule(lr){9-10} 
                        & B-1  & B-2  & B-3  & B-4  & R-1      & R-2    & R-3    & D-1           & D-2           & \multicolumn{1}{c}{}                        \\ \midrule
BoB             & 4.69  & 1.57 & 0.65 & 0.31 & 10.68   & 1.57   & 9.30   & 4.94         & 17.06         & 3.97                                        \\ \midrule
w/o. \textit{know}               & 6.37 & 2.13 & 1.07 & 0.69 & 9.68     & 1.41   & 8.06   & 7.19          & 25.87         & 3.87                                        \\
w/o. \textit{mem}               & 6.79 & 1.90  & 0.65 & 0.23 & 9.79     & 1.16   & 8.11   & 5.17          & 16.95         & 3.91                                        \\
w/o. \textit{dual}               & 8.58 & 3.74 & 2.42 & 1.84 & 12.05    & 2.80   & 9.87   & 8.74          & 34.23         & 4.97      \\
w/o. \textit{dep}               & 8.64 & 3.29 & 1.90  & 1.35 & 10.78    & 1.96   & 8.65   & 9.03          & 36.01         & 4.57                                        \\

\thickhline
\end{tabular}
\caption{Ablation results \textbf{(RQ1)}.}
\vspace{-4mm}
\label{tab:ablation}
\end{table*}

\subsection{Evaluation Metrics}
We choose distinctness, BLEU\cite{papineni2002bleu}, ROUGE\cite{lin2004rouge}\footnote{\scriptsize\url{https://github.com/bckim92/language-evaluation}} and METEOR\cite{denkowski:lavie:meteor-wmt:2014}\footnote{\scriptsize\url{https://github.com/Maluuba/nlg-eval}} to be our automatic metrics.
Focusing on the exact n-gram co-occurrence in hypothesis and reference, BLEU and ROUGE evaluate the appropriateness of the proposed model. Distinctness is calculated as the ratio of unique unigrams and bigrams, paying more attention to the diversity of generated text.  METEOR measures the alignment, or the exact, stem, synonym, and paraphrase matches between the hypothesis and reference.

Apart from automatic evaluation, we conduct human evaluation. Specifically, $200$ examples are randomly sampled from the test set and well-educated native speakers are recruited to assess the quality of the generation from different models with their source hidden. Each annotators are required to give a score in $\{0: \mathrm{bad},1:  \mathrm{fair}, 2:\mathrm{good}\}$  for three independent aspects: (1) \textit{fluency}: whether the reply is fluent; (2) \textit{coherence}: whether the reply is coherent with the context; and (3) \textit{faithfulness}: whether the reply is well-grounded and faithful to the selected knowledge sentence and memory fragment. The agreement of annotators is measured via Fleiss' kappa~\cite{fleiss1971measuring}.

\subsection{Experiment Results}
We first report the experimental result in automatic evaluation.
As is shown in Table~\ref{tab:main_exp}, our method outperforms the state-of-the-art baselines in KGC and personalized dialogue in most metrics, verifying the effectiveness of our model empirically. 
Among non-pretrained methods, TMN and GPMN are low in diversity, since their generator is not pre-trained on large corpus before. SKT improves distinctness but shows low appropriateness, possibly because that it highly relies on the golden knowledge label, which is costly and not always available.
In pre-trained based methods, Transfertransfo attains impressive results on distinctness. It also achieves competitive appropriateness results, but not as good as ours. We gauge the performance of the model to the large document-level training corpus, a critical choice for pre-trained language model, which may boost the diversity of generated text. 
Besides, the performance of the BoB, a recently published baseline, is less satisfactory compared with others. The premise of BoB is the disentanglement between contextual coherence and persona consistency, which is not always achievable especially when we use user-specific dialogue history for personal memory information.
And it is notable from the table that there is a significant gap between the baseline methods in KGC or personalized dialogue and ours, validating that neither simply projecting personal information into dialogue nor purely grounding on knowledge is an acceptable solution to the KGC task. It is necessary to combine personal memory and external knowledge together.  
The comprehensive improvement of KnowledGPT+M in contrast with the original KnowledGPT also reveals this viewpoint. Additionally, the considerable advantage of our proposed method over KnowledGPT+M illustrates the fact that treating personal memory as knowledge is not enough. The dependency between personal memory and the knowledge should not be ignored.

We also present the result of human evaluation since no automatic metric is perfect in this task~\cite{dinan2018wizard}. Since human evaluation is time-consuming and expensive, only competitive baselines are involved. As shown in Table~\ref{tab: human}, our proposed model outperforms the baseline methods and there is an evident improvement.

\begin{figure}[!t]
    \centering
    \vspace{-3mm}
    \includegraphics[width=0.8\linewidth]{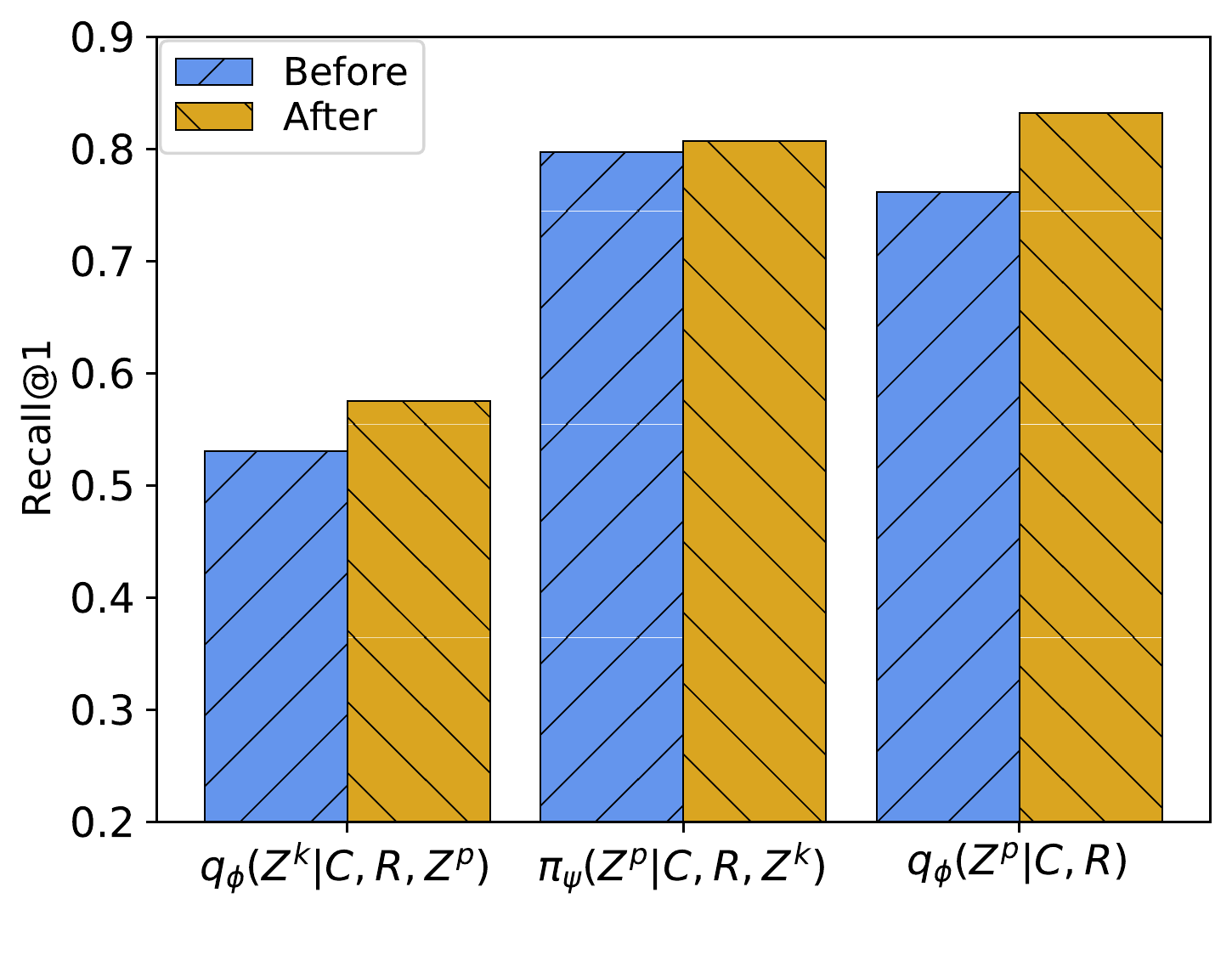}
    \vspace{-3mm}
    \caption{The Recall@1 of knowledge (or personal memory) before and after the closed dual loop.}
\vspace{-5mm}
    \label{fig:dual}
\end{figure}

\begin{figure}[!t]
    \centering
    \includegraphics[width=0.85\linewidth]{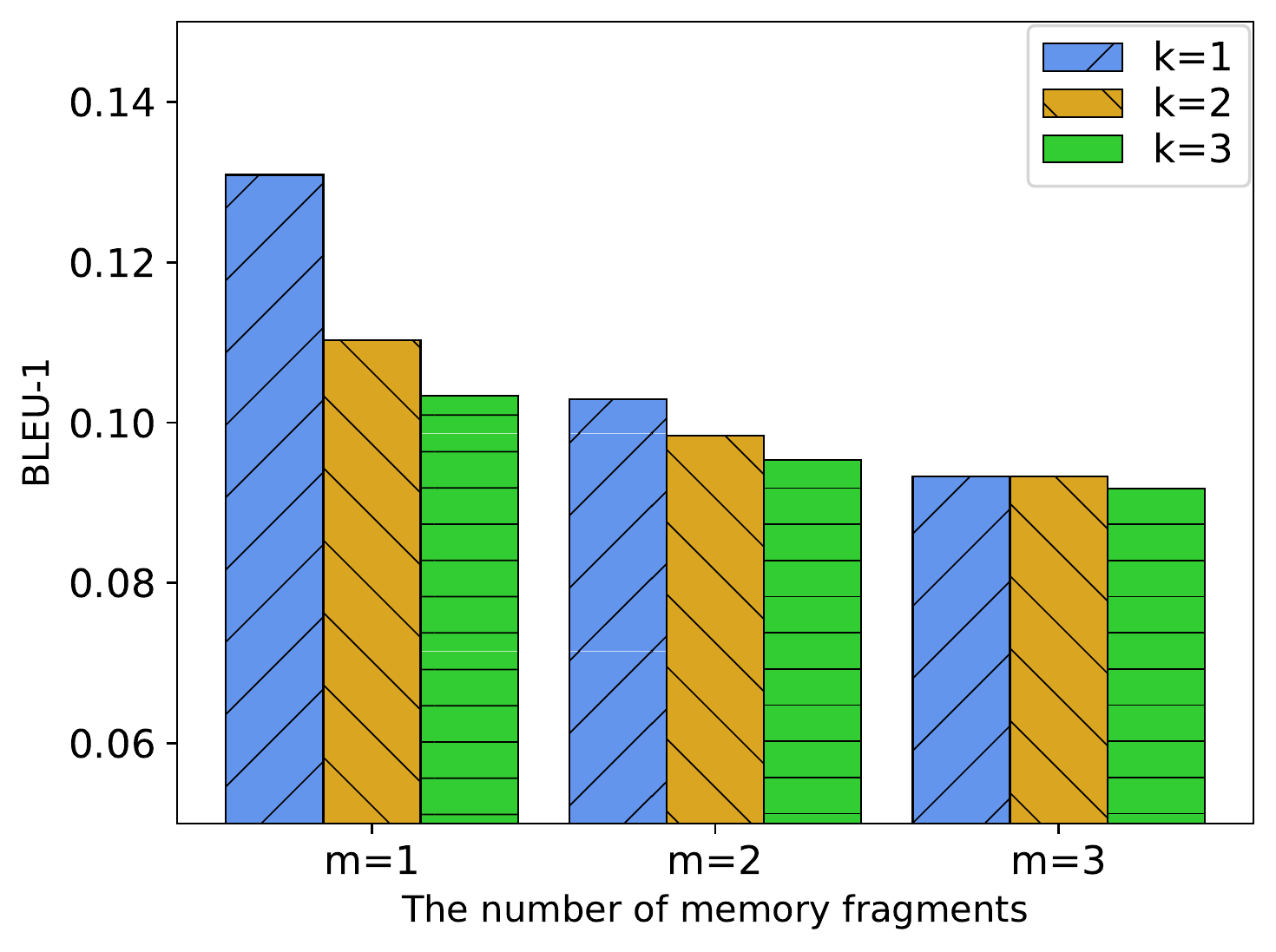}
    \vspace{-1mm}
    \caption{The performance of our model in terms of BLEU-1 under different number of personal memory fragment and knowledge.}
    \label{fig:m}
\vspace{-2mm}
\end{figure}

\subsection{Analysis}
Apart from the main results, we are especially interested in some research questions:
\begin{itemize}[noitemsep,topsep=0.4pt,parsep=0pt,partopsep=0pt]
    \item \textbf{(RQ1)} How does each component contributes to the performance of our model?
    \item \textbf{(RQ2)} How many knowledge sentences and memory fragments to select?
\end{itemize}
  
To answer the first question, we conduct ablation study and compare the full model with several variants:(1) w/o. \textit{know}. the external knowledge base to grounding the dialogue is removed; (2) w/o. \textit{mem}.  personal memory is removed and this variant is a standard KGC model essentially; (3) w/o. \textit{dual}. the dual task is removed, so there is no dual learning and distillation in this variant; (4) w/o. \textit{dep}. the dependency of the two latent variables is removed so $Z^p$ and $Z^k$ are calculated independently.
The ablation result is shown in Table~\ref{tab:ablation}, from which we could have the following observations:
(1) w/o. \textit{know} and w/o. \textit{mem} exhibit a degeneration at a great extent, further justifying the necessity of introducing knowledge and personal memory into a dialogue system, respectively. 
(2) w/o. \textit{dep} also shows an obvious deterioration. This is in line with our expectation since w/o. \textit{dep} model $Z^k$ and $Z^p$ as two independent latent variables, ignoring the underlying dependence between them. Comparatively speaking, w/o. \textit{dual} achieves a better result, but not as good as the full model due to the destroy of the closed dual loop.

And to have a intuitive perception about the effect of the closed dual loop, we examine the promotion brought to the $q_\phi(Z^k|C,R,Z^p)$, $\pi_\psi(Z^p|C,R,Z^k)$ and $q_\phi(Z^p|C,R)$ in terms of Recall@1 of knowledge or personal memory. The result is shown in Figure~\ref{fig:dual}. From the figure we could see that there is an obvious improvement after trained with our proposed learning algorithm.

For the \textbf{(RQ2)}, we first explore it by varying the amount of selected personal memory fragments and observe how the knowledge selection procedure is influenced. In detail, we vary the number of personal memory fragments $m$ sampled by $p_\theta(Z^p|C)$ from $1$ to $4$ and evaluate the performance of $p_\theta(Z^k|C,Z^p)$ in terms of Recall@n (n$\in$\{1,2,5,10\}).

\begin{table}[]
\resizebox{\linewidth}{!}{
\begin{tabular}{ccccc}
\toprule
    & Recall@1 & Recall@2 & Recall@5 & Recall@10 \\
    \midrule
m=1 & 0.173    & 0.286    & 0.505    & 0.720     \\
m=2 & 0.176    & \textbf{0.289}    & \textbf{0.513}    & 0.730     \\
m=3 & \textbf{0.177}    & 0.289    & 0.509    & 0.730     \\
m=4 & 0.176    & 0.288    & 0.508    & 0.730    \\
\bottomrule
\end{tabular}
}
\caption{The performance of $p_\theta(Z^k|C,Z^p)$ under different $m$. The numbers in bold are the best results.}
\label{tab: vae}
\vspace{-4mm}
\end{table}

As is shown in Table~\ref{tab: vae}, we could find that the best performance is reached when $m=2$. There is a fluctuation or slight drop when $m$ continues to increase possibly owing to the distraction mixed with the redundant personal memory.
Besides, we are also curious about the final generation performance under different numbers of knowledge and personal memory fragment. It could be seen from Figure~\ref{fig:m} that there appears a decline when we increase the number of knowledge and personal memory fragment, which we attribute to the unwanted noise mixed with personal memory and knowledge.
\vspace{-1.5mm}
\section{Conclusion}
\label{sec:conclusion}
\vspace{-1mm}
In this work, we explore personalized KGC by introducing personal memory into knowledge selection task. Two latent variables are introduced to select knowledge and personal memory respectively. Besides, dual learning scheme is employed to allow the two selection task to teach each other. For future work, we would like to extend the personalized knowledge-grounded dialogue to personalized conversational recommendation system for application in online shopping.

\section*{Ethical Considerations}

\paragraph{Intended Use} The chief purpose of our dataset is to examine a  dialogue model's capacity in selecting proper knowledge with the help of personal memory. The dataset is mainly for research propose and it is not supposed to be directly used to train a production system. And researchers should be aware of the possible ethic issues before exploiting our dataset. 

\paragraph{Data Collection} All the examples in our dataset are in English and no human annotators are involved in the data collection process. As mentioned in Sec.\ref{sec:dataset}, our dataset is built on the basis of the Reddit dumps from Pushshift~\cite{baumgartner2020pushshift}, which is a publicly available resource widely used in more than a hundred peer-reviewed publications. Our data collection is in consistent with the term of use and the research is granted ethical approval by an external institutional review board.  To avoid potential abuse, the dataset is available upon request to the authors. Contact the authors (by email) and clearly state your intended use if you believe the dataset might be helpful in your research.

\paragraph{User Privacy} Although our dataset includes user-specific utterance history as personal memory, no user account names will be revealed or inferred from the dataset. Besides, the utterance histories are paraphrased during our procession of the dataset such that they can not be traced back to the real users in Reddit. In conclusion, There is no \textit{personally identifiable information} in our dataset or underlying leakage of personal information. 

\section*{Acknowledgement}
Thanks for the reviewers for their valuable suggestions. This work was supported by National Natural Science Foundation of China (NSFC Grant No. 62122089 \& No. 61876196 \& No. 61832017), Beijing Outstanding Young Scientist Program NO. BJJWZYJH012019100020098, and Intelligent Social Governance Platform, Major Innovation \& Planning Interdisciplinary Platform for the "Double-First Class" Initiative, Renmin University of China. We also wish to acknowledge the supports provided and contributions made by Public Policy and Decision-making Research Lab of RUC, and the Public Computing Cloud, Renmin University of China. Rui Yan is also supported by Beijing Academy of Artificial Intelligence~(BAAI).

\bibliography{custom}
\bibliographystyle{acl_natbib}

\clearpage
\appendix

\section{Appendix}

\subsection{The derivation of ELBO}
\label{sec:appendix}

\begin{small}
\begin{equation}
\begin{aligned}
&\mathcal{L}_{elbo}=\log p(R)\\
&=\log \int_{Z^k}\int_{Z^p} p(R, Z^p,Z^k) dZ^pdZ^k \\
&=\log \int_{Z^k}\int_{Z^p} g(R|Z^p,Z^k) p(Z^p,Z^k) dZ^pdZ^k \\
&=\log \int_{Z^k}\int_{Z^p} g(R|Z^p,Z^k) p_\theta(Z^k|Z^p) p_\theta(Z^p) dZ^pdZ^k\\
&=\log \int_{Z^k}\int_{Z^p} g(R|Z^p,Z^k) p_\theta(Z^k|Z^p) p_\theta(Z^p)\\
& \cdot \frac{q(Z^p,Z^k)}{q(Z^p,Z^k)}dZ^pdZ^k\\
&=\log \int_{Z^k}\int_{Z^p} g(R|Z^p,Z^k) p_\theta(Z^k|Z^p) p_\theta(Z^p) \\
& \cdot \frac{q(Z^k|Z^p)q(Z^p)}{q(Z^k|Z^p)q(Z^p)}dZ^pdZ^k\\
&=\log \mathbb{E}_{q(Z^k|Z^p)q(Z^p)} g(R|Z^p,Z^k) \frac{p_\theta(Z^k|Z^p) p_\theta(Z^p)}{q(Z^k|Z^p)q(Z^p)} \\
&\geq \mathbb{E}_{q(Z^k|Z^p)q(Z^p)} \log  g(R|Z^p,Z^k) \frac{p_\theta(Z^k|Z^p) p_\theta(Z^p)}{q(Z^k|Z^p)q(Z^p)} \\
&= \mathbb{E}_{q(Z^k|Z^p)q(Z^p)} \log g(R|Z^p,Z^k)\\ 
&+\mathbb{E}_{q(Z^k|Z^p)q(Z^p)} [\log p_\theta(Z^k|Z^p)-\log q(Z^k|Z^p)]\\
&+\mathbb{E}_{q(Z^k|Z^p)q(Z^p)} [\log p_\theta(Z^p)-\log q(Z^p)]\\ 
\end{aligned}
\end{equation}
\end{small}

For the first term, it could be decomposed as:
\begin{equation}
\begin{aligned}
& \mathbb{E}_{q(Z^k|Z^p)q(Z^p)} \log g(R|Z^p,Z^k)\\
&=\mathbb{E}_{q(Z^k|Z^p)q(Z^p)} \sum\limits_{i=1}^{l_R} \log g(R|Z^p,Z^k,r_{<i}) 
\end{aligned}
\end{equation}

For the second term and the third term, they could be further simplified:
\begin{equation}
\begin{aligned}
&\mathbb{E}_{q(Z^k|Z^p)q(Z^p)} [\log p_\theta(Z^p)-\log q(Z^p)]\\
&=-KL(q_\phi(Z^p)||p_\theta(Z^p))
\end{aligned}
\end{equation}

\begin{equation}
\begin{aligned}
&\mathbb{E}_{q(Z^k|Z^p)q(Z^p)}[\log p_\theta(Z^k|Z^p)-\log q(Z^k|Z^p)]\\
&=-\mathbb{E}_{q_\phi(Z^p)}KL(q_\phi(Z^k|Z^p)||p_\theta(Z^k|Z^p))\\
\end{aligned}
\end{equation}

\subsection{Implementation Details}
\label{sec:impl}

We choose BERT$_{base}$~\cite{devlin2018bert}\footnote{\scriptsize\url{https://huggingface.co/bert-base-uncased}} and GPT-2~\cite{radford2019language}\footnote{\scriptsize\url{https://huggingface.co/gpt2}} as the pre-trained language model, and implement our methods with the code in Hugging Face. 
To tag the pseudo knowledge label and personal memory label, the similarity score function used in Eq.~\ref{eq:pseudo} is implemented as unigram F1~\cite{dinan2018wizard} with the code shared at ParlAI \footnote{
\scriptsize\url{https://github.com/facebookresearch/ParlAI/blob/master/parlai/core/metrics.py}}.
In the warm up phase, we pre-train the primal task and dual task for $5000$ steps and set the batch size and learning rate to be $16$ and $1e-5$ respectively. The posterior distribution of $Z^p$ is optimized for 1000 steps with a learning rate of $1e-5$ and a batch size of $16$. 
In the dual learning phase, the algorithm~\ref{alg:alg} runs for $1000$ steps with a batch size of $16$ and a learning rate of $1e-6$.
All modules are learned with Adam on a GTX 1080, and we set the hyperparameter of Adam to be  $\beta_1=0.9$, $\beta_2=0.999$ respectively. Cosine learning schedule is applied to  adjust the learning rate during training. We set the minimum learning rate to be $0$ in cosine learning schedule. Gradient clip is set to $2.0$ to avoid the explosion of gradient. 
When decoding, beam search is applied with a beam width of $5$ and the minimum generated length is $10$. The repetition penalty and the length penalty is set to be $1.0$ and $0.0$ respectively.

\subsection{Data Examples}
\label{sec:example}
In Table~\ref{tab:example}, We present several examples of our constructed dataset.

\begin{table}
    \centering
    \includegraphics[width=1\linewidth]{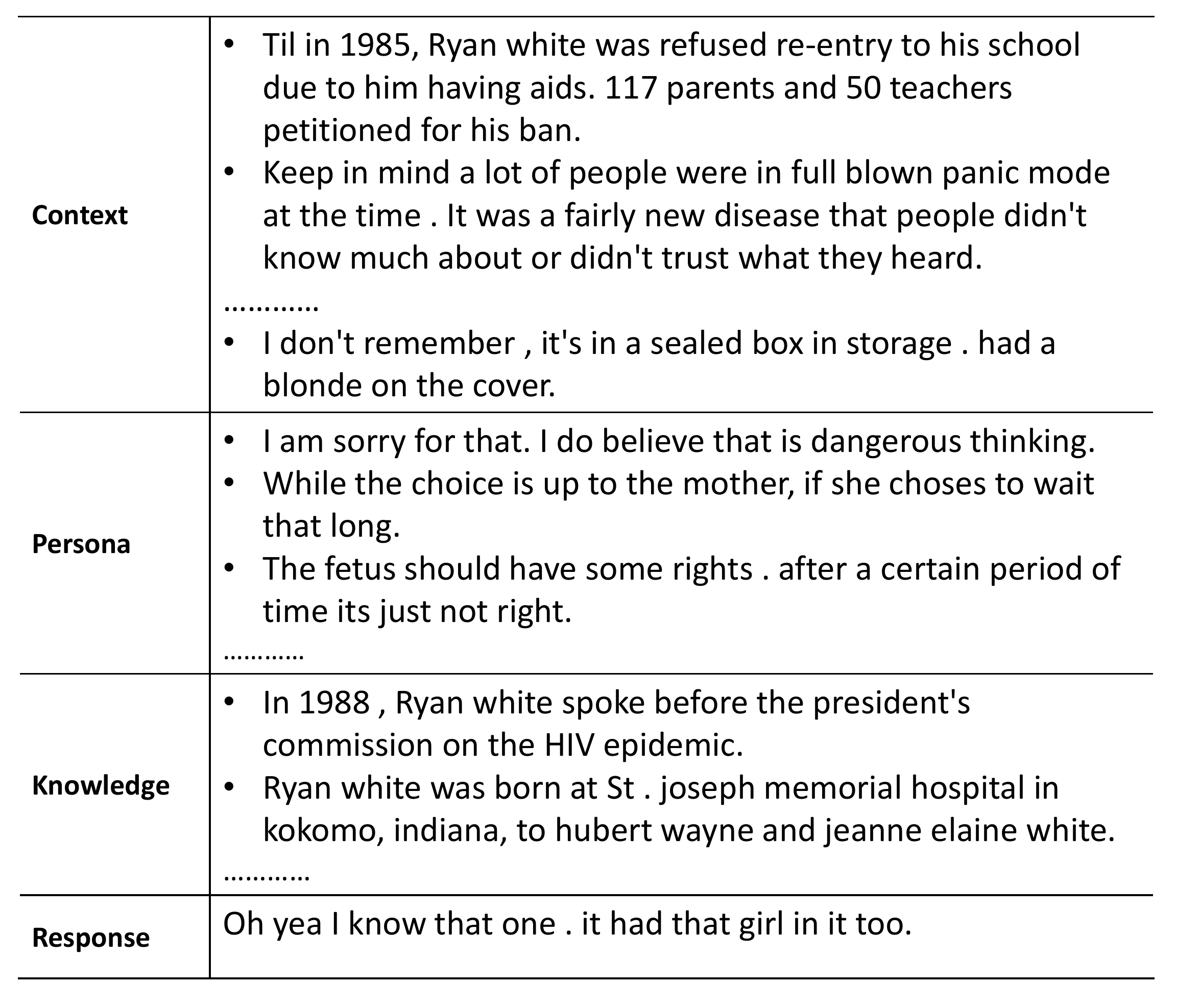}
    \includegraphics[width=1\linewidth]{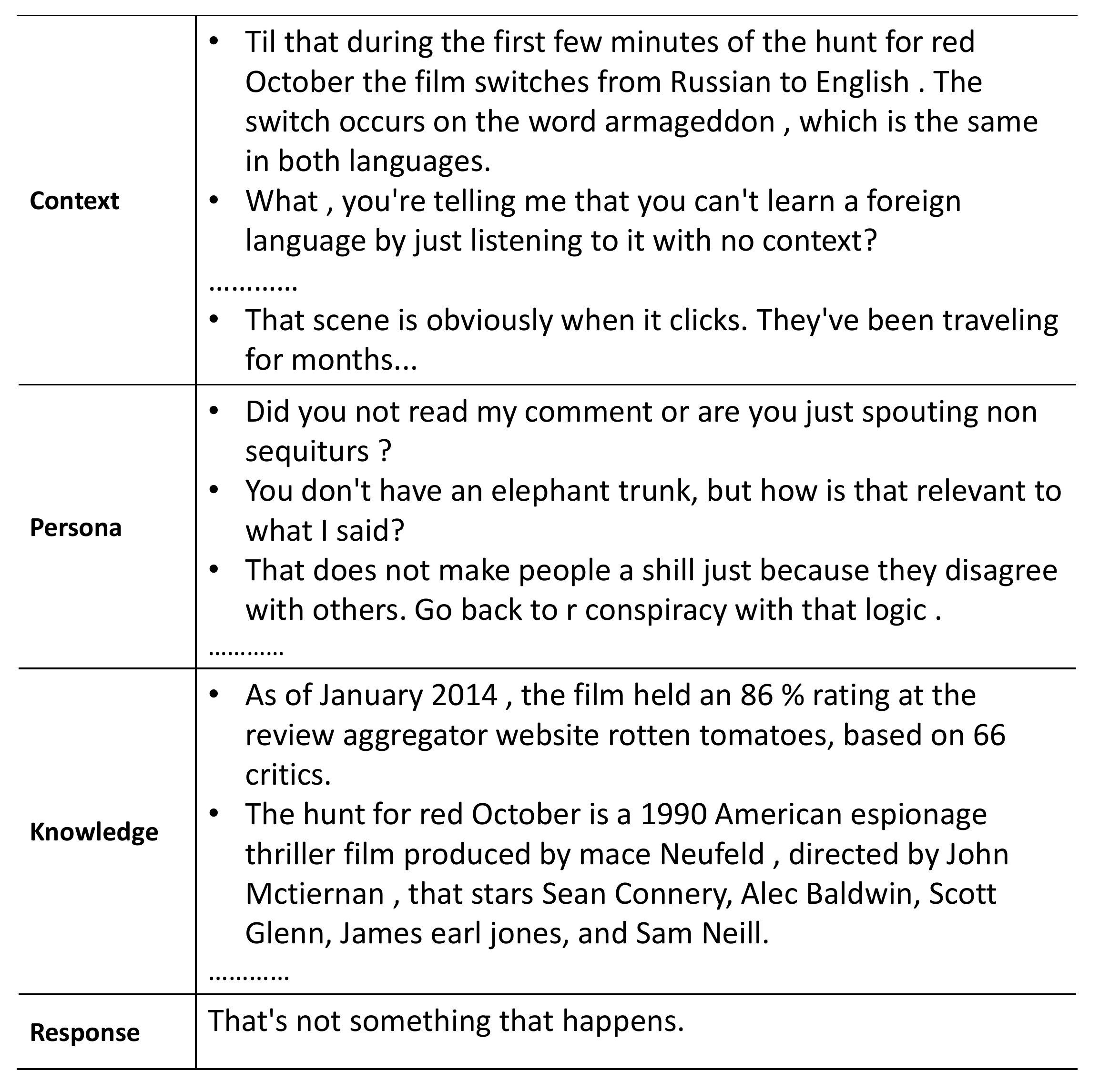}
    \caption{Examples of our constructed dataset.}
    
    \label{tab:example}
\vspace{-7mm}
\end{table}

\subsection{Case Study}
\label{sec:case}
To further analyse the model's features, a case in test set is provided in Table~\ref{tab:case}. As is shown, baseline methods in personalized dialogue has no access to external knowledge and facts, thus their generation result tend to be a little generic. And it seems that the ordinary KGC methods usually give a plain response like KnowledGPT. Our proposed method generates a more human-like response, which is in line with our expectation. 

\begin{table}[!h]
\resizebox{1.0\linewidth}{!}{
\begin{tabular}{p{2.5cm}<{\raggedleft} p{10.5cm}<{\raggedright}}
\toprule
\multicolumn{2}{p{13cm}<{\centering}}{\textbf{\emph{Knowledge}}} \\ \midrule
\multicolumn{2}{p{13cm}<{\raggedright}}{$\bullet$ ...} \\
\multicolumn{2}{p{13cm}<{\raggedright}}{$\bullet$ Advertisement the recent experiment, however, addressed this concern head-on, while also demonstrating the engine's potential to work in space. }\\
\multicolumn{2}{p{13cm}<{\raggedright}}{$\bullet$ He em drive's thrust was due to the quantum vacuum behaving like propellant ions behave in a magnet ohydrodynamics drive for spacecraft propulsion.} \\
\multicolumn{2}{p{13cm}<{\raggedright}}{$\bullet$ Advertisement serious inquiry, indeed. it's crucial now that these tests be analyzed, replicated, and confirmed elsewhere. A peer-review and formal paper would also seem to be in order lest we get too carried away with these results. But wow. Just wow. }\\
\multicolumn{2}{p{13cm}<{\raggedright}}{$\bullet$ It's still early days, but the implications are mind-boggling to say the least. A full-fledged em drive could be used on everything from satellites working in low earth orbit, to missions to the moon, mars, and the outer solar system. }\\
\multicolumn{2}{p{13cm}<{\raggedright}}{$\bullet$ ...}  \\ \midrule
\multicolumn{2}{c}{\textbf{\emph{Personal Memory}}} \\  \midrule
\multicolumn{2}{p{13cm}<{\raggedright}}{$\bullet$ ...} \\ 
\multicolumn{2}{p{13cm}<{\raggedright}}{$\bullet$Frankly, i'd expect the constitution to win in a fight against a north korean submarine.
 } \\ 
\multicolumn{2}{p{13cm}<{\raggedright}}{$\bullet$Modern diesel subs can be pretty spectacular. The israeli dolphin class are top notch, and they can be quieter than nuclear submarines if they want to be. } \\
\multicolumn{2}{p{13cm}<{\raggedright}}{$\bullet$Russia'll sell new kilo class subs to pretty much anyone who'll pay.} \\
\multicolumn{2}{p{13cm}<{\raggedright}}{$\bullet$ ...}  \\ \midrule

\multicolumn{2}{c}{\textbf{\emph{Context}}} \\ \midrule
\textbf{U1:} &  New test suggests nasa's impossible em drive will work in space the em appears to violate conventional physics and the law of conservation of momentum the engine converts electric power to thrust without the need for any propellant by bouncing microwaves within a closed container. \\
\textbf{U2:} &  In fairness it usually turns out to be false . \\
\textbf{U3:} &  It's an extraordinary claim, and we don't have extraordinary proof yet. It'll be a while before we have a yes no on the emdrive. \\

\midrule
\multicolumn{2}{c}{\textbf{\emph{Response}}} \\ \midrule
\multirow{1}{*}{GPMN}& I't have a lot of the same point view.\\
\midrule
\multirow{1}{*}{TMN} & I'm not sure it's not a lot of people convinced by that.\\
\midrule
\multirow{1}{*}{Transfertransfo} & The police would a be a better case scenario, as the officers the people claiming to be the best cops would have police. \\
\midrule
\multirow{1}{*}{SKT}& What the fuck is wrong with the fact that the guy is trying to prove to have extraordinary proof? \\
 \midrule
\multirow{1}{*}{KnowledGPT} & It generates thrust when it's in a vacuum. \\
\midrule 
\multirow{1}{*}{KnowledGPT+M} & it's not rocket science , it's physics . \\
\midrule
\multirow{1}{*}{BoB} & i'm not saying it can be done, just that it can be done in a way \\
\midrule
\multirow{1}{*}{$\mathcal{P}^2$BOT} & it's not like we're going to be able to get rid of this.\\
\midrule
\multirow{1}{*}{Ours} & Yes, but it's not a scientific breakthrough. It's an extraordinary claim, and we don't have extraordinary proof yet. \\
\midrule
\multirow{1}{*}{Human} & No one reputable is willing yet to rule out experimental error of some sort. The vacuum test rules out one of the outstanding possibilities, but it's by no means the final word. as the announcement says, the question of where the thrust is coming from deserves serious inquiry. \\

\bottomrule

\end{tabular}
}
\caption{A case from the test set.}
\label{tab:case}

\end{table}

\end{document}